%% file: paper.tex
\documentclass{llncs}
\usepackage{makeidx}  

\input{packages}
\input{commands}

\begin{document}
\frontmatter          
\pagestyle{headings}  
%
\mainmatter              
\title{Minimum Displacement Motion Planning for Movable Obstacles}
%
%
\author{Antony Thomas \and Fulvio Mastrogiovanni}
\authorrunning{Antony Thomas and Fulvio Mastrogiovanni} 
%

%
\institute{Department of Informatics, Bioengineering, Robotics, and Systems Engineering, University of Genoa, Via All'Opera Pia 13, 16145 Genoa, Italy.\\
\email{antony.thomas@dibris.unige.it, fulvio.mastrogiovanni@unige.it}}

\maketitle              

\begin{abstract}
This paper presents a minimum displacement motion planning problem wherein obstacles are displaced by a minimum amount to find a feasible path. We define a metric for robot-obstacle intersection that measures the extent of the intersection and use this to penalize robot-obstacle overlaps. Employing the actual robot dynamics, the planner first finds a path through the obstacles that minimizes the robot-obstacle intersections. The metric is then used to iteratively displace the obstacles to achieve a feasible path. Several examples are provided that successfully demonstrates the proposed problem.  
\end{abstract}

\section{Introduction}
\input{introduction}

\section{Minimum Displacement Motion Planning}
\input{mdmp}

\section{Algorithm for Minimum Displacement Motion Planning}
\label{sec:approach}
\input{approach}

\section{Examples and Applications}
\input{examples}
\section{Related Work}
\input{related}
\section{Conclusion}
\input{conclusion}
\bibliographystyle{splncs03.bst}
\bibliography{References}
\end{document}

%% file: packages.tex
\usepackage{url}
\usepackage{makeidx}         
\usepackage{graphicx}        
\usepackage{multicol}        
\usepackage[bottom]{footmisc}

\usepackage{amsmath,bm,amsfonts,amssymb}

\usepackage{commath}      
\usepackage{color,xcolor,ucs}
\usepackage{adjustbox}
\usepackage{subfig}
\usepackage[font=small,labelfont=bf]{caption}

\usepackage{floatrow}
\usepackage{tabularx}
\usepackage{float}
\usepackage{booktabs}
\usepackage{cite}
\usepackage{xr-hyper}
\usepackage{wrapfig}

\usepackage{algorithm}
\usepackage{algpseudocode}
\usepackage{multirow}
\usepackage{listings}
\usepackage{lipsum}
\usepackage{textcomp} 

\usepackage{array}
\usepackage[nocomma]{optidef}  

%% file: commands.tex
\makeatletter
\newcommand{\clearsubcaptcounter}{\setcounter{sub\@captype}{0}}
\makeatother

\newcolumntype{C}[1]{>{\centering\arraybackslash}p{#1}}

%% file: introduction.tex
Classical robot motion planning problem finds a collision free path from a given initial pose to a desired goal pose of a robot amidst fixed obstacles~\cite{latombe1991robot}. This problem is well studied. However, in several situations a robot may need to move all (if no collision free path exists) or certain obstacles along its path. Mobile robots in a indoor environment may need to move obstacles to reach a goal. For example, open closed doors or push chairs around to reach a location. In the case of dynamic obstacles a robot may need to plan a path through these obstacles as they might clear a path while moving. Manipulator robots often rearrange or move obstacles aside to complete a task. 

\begin{figure}[t]
  \subfloat[]{\includegraphics[trim=52 110 38 101,clip,scale=0.5]{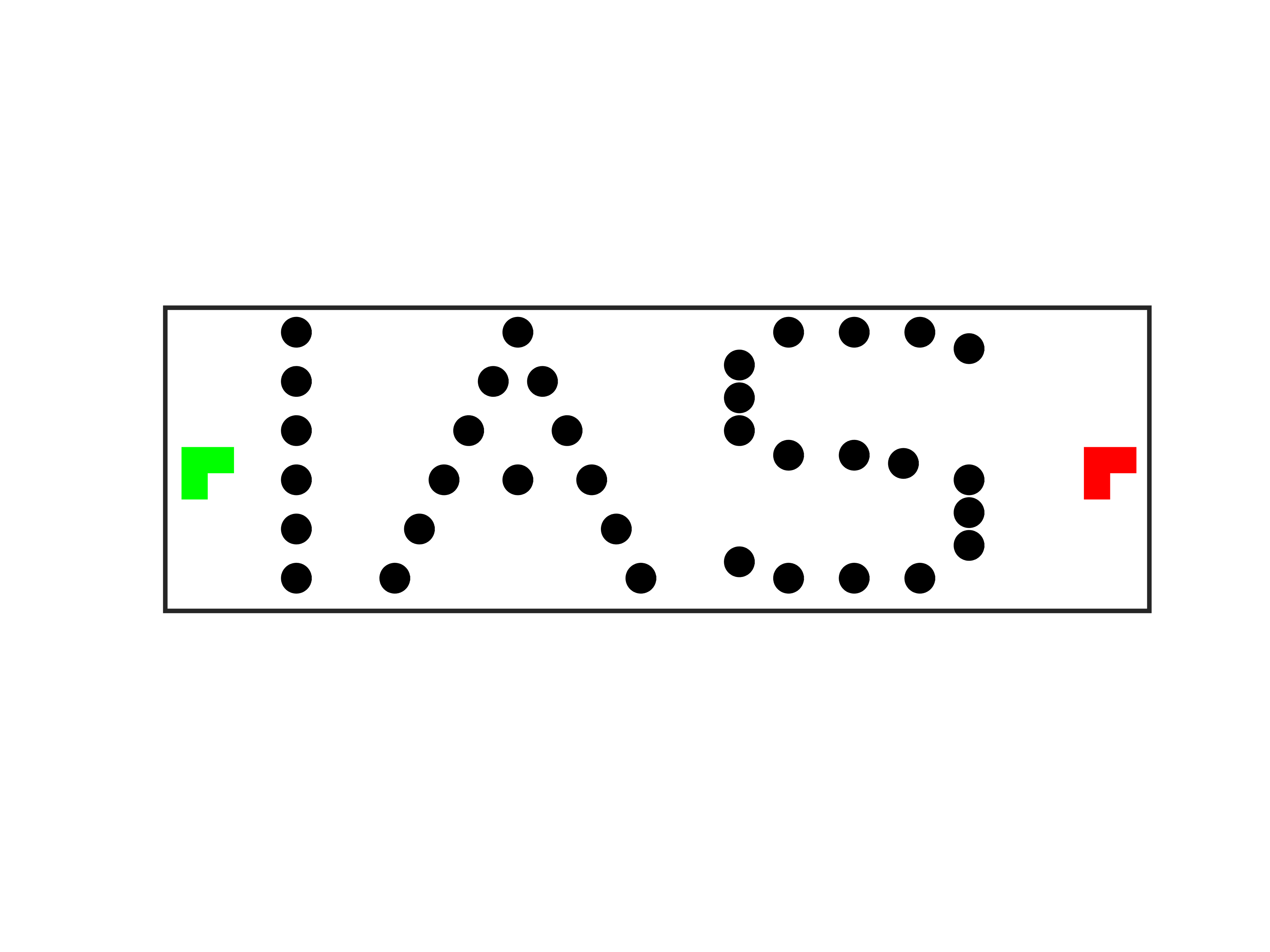}\label{fig:L1}}\hfill
  \subfloat[]{\includegraphics[trim=52 110 38 101,clip,scale=0.5]{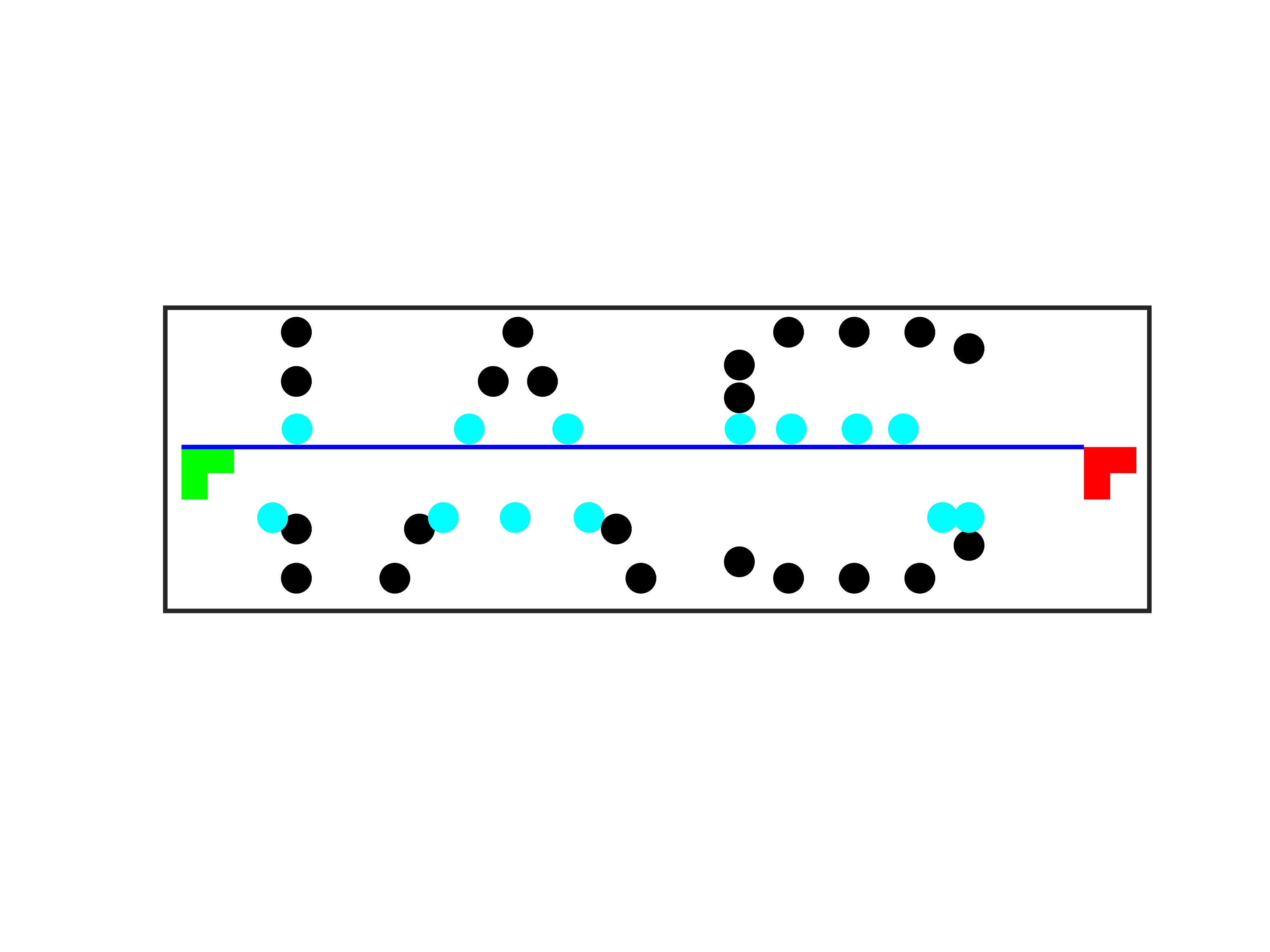}\label{fig:L2}}\hfill
  \subfloat[]{\includegraphics[trim=52 110 38 101,clip,scale=0.5]{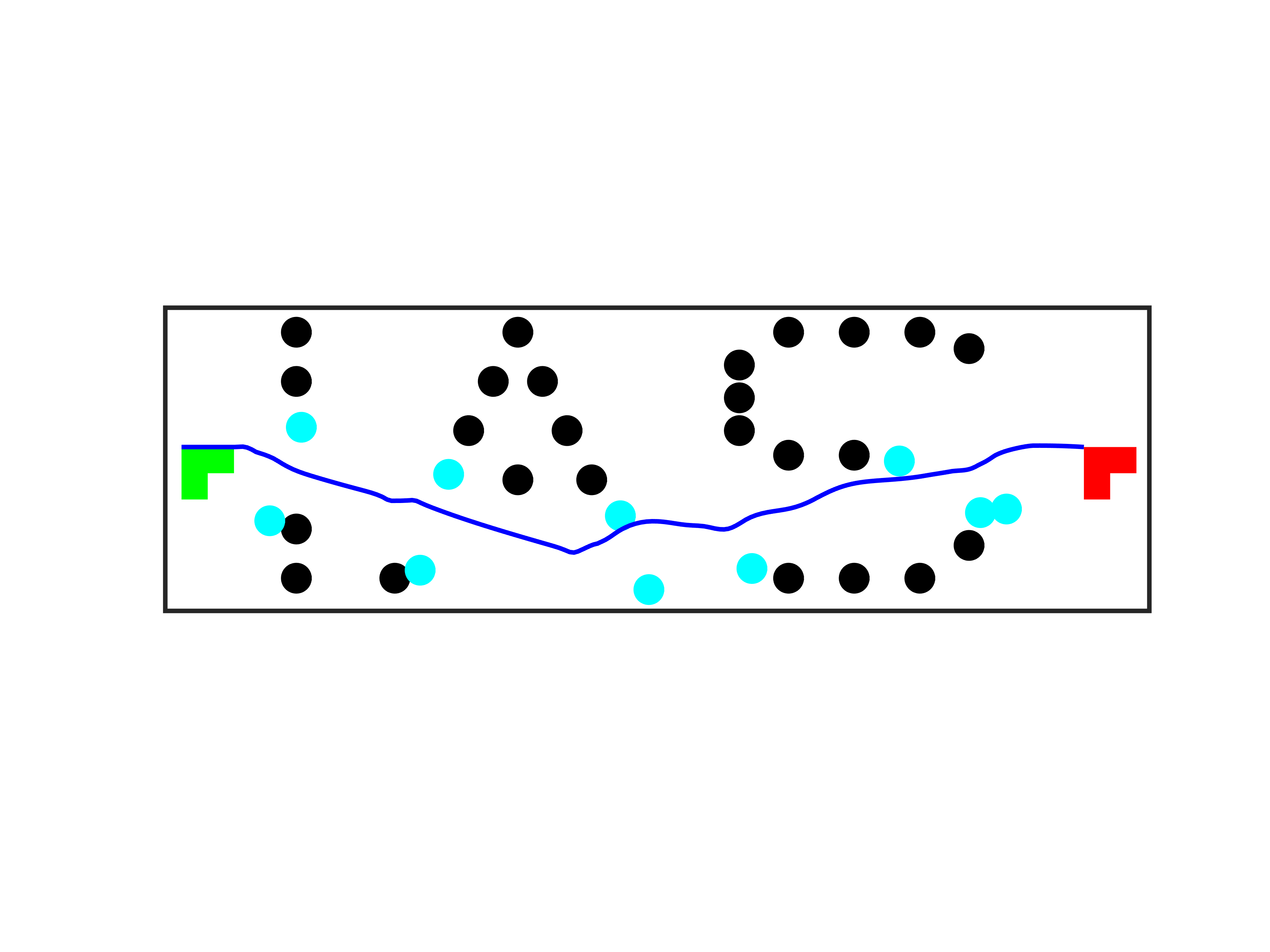}\label{fig:L3}}\hfill
  \subfloat[]{\includegraphics[trim=52 110 38 101,clip,scale=0.5]{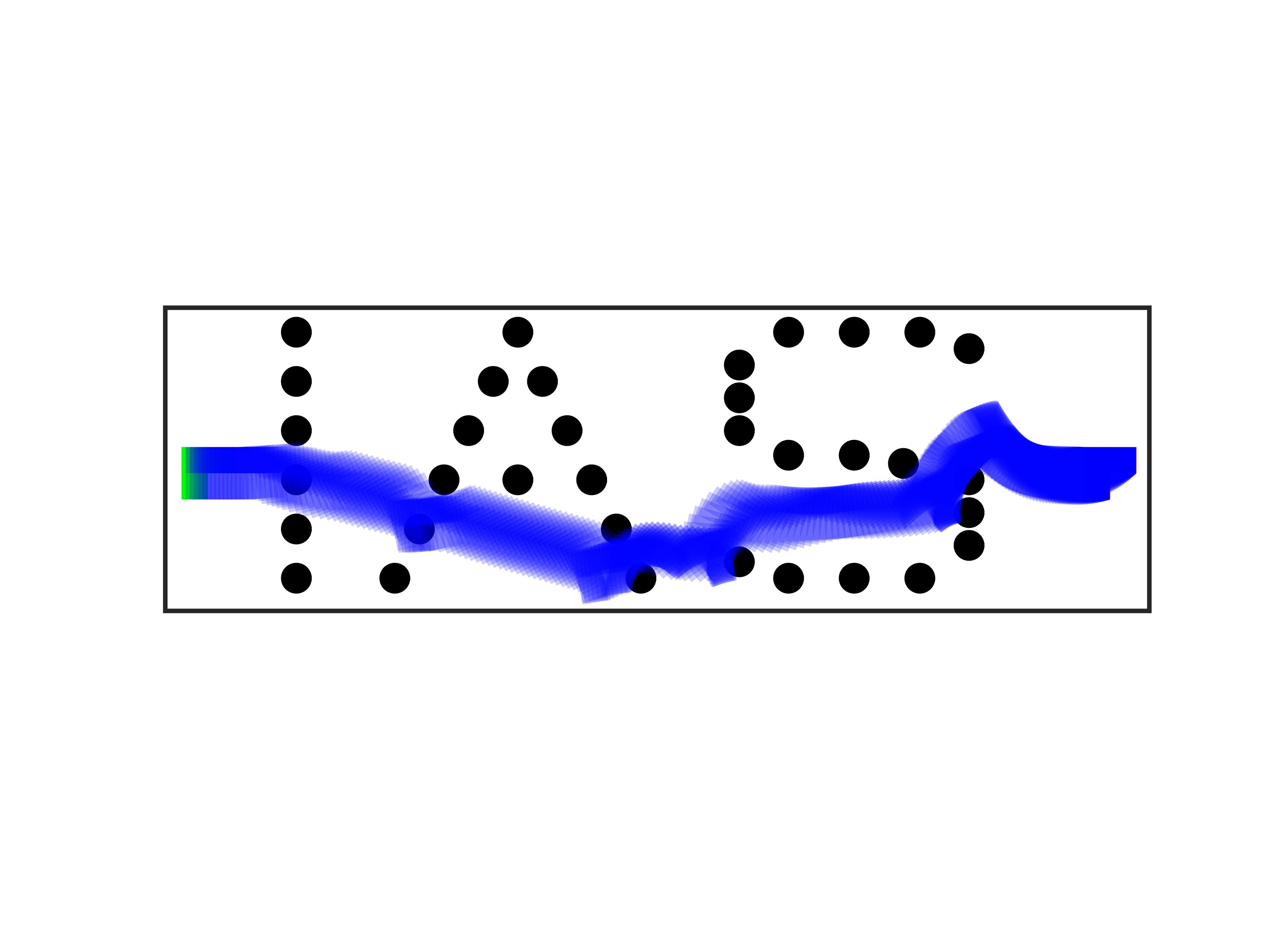}\label{fig:L4}}\hfill
   \caption{(a) An L-shaped robot in the \textit{IAS domain} has to move from left to right by displacing obstacles. (b) Robot takes the shortest path and the 13 displaced obstacles can be seen in cyan color. (c) Robot finds a path with smaller displacement, moving 10 obstacles. (d) An example of the planning phase of our approach where a path though the obstacles is obtained that minimizes the robot-obstacle overlaps.}
  \label{fig:L}
\end{figure}

Such problems require the knowledge about the environment and the robot's motion. A collision between a robot and an obstacle occur when they intersect each other. If the robot has to clear its way to find a feasible path then the obstacle must be displaced such that there is no robot-obstacle intersection. In this paper, we investigate this problem, that is, finding a feasible path such that the obstacle displacements are minimized (see Fig.~\ref{fig:L}). Currently we do not focus on how the displacement happens and assume that the computed displacements can be achieved irrespective of the type of robot-obstacle interaction required to move the obstacles.  

We present a new motion planning formulation in which the planner first finds a path through the obstacles such that the robot-obstacle intersections are minimized. To this end, the planner minimizes a weighted sum of the robot path length and a metric that measures the extent of robot-obstacle intersection. This robot-obstacle intersection metric is then used to determine the minimum displacements for the obstacles so that the computed path is collision free. While computing the minimum displacements, we also take into account the fact that the displaced obstacles must themselves be collision free.  

%% file: mdmp.tex
Let $\mathcal{C}$ denote the robot configuration space with $x^0,  x^g \in \mathcal{C}$ denoting the start and goal states. Let $\mathcal{O}= \{o^i| \ 1 \leq i \leq n\}$ denote the set of movable obstacles in the environment with an associated displacement set $\mathcal{D}= \{d^i| \ 1 \leq i \leq n\}$ that represents the corresponding obstacle displacements. By abuse of notation we will use $o^i$ to denote both the $i$-th obstacle as well as its state. Note that the displacements $d^i$ in general may not always be translational alone and can include rotations or deformations. Thus $d^i$ belongs to a displacement space of arbitrary dimension.  In this paper, we will consider obstacles that can either perform translation alone or pure rotation alone about its center. Therefore $d^i$ either represents pure translation or pure rotation. Initially, $d^i = 0$. We can now define the minimum displacement motion planning problem.
\begin{definition}
The minimum displacement motion planning problem from $x^0$ to $x^g$ finds a path $x: [0,1] \rightarrow \mathcal{C}$ with obstacle displacements $d^1,\ldots,d^n$ minimizing the cost $C(x,\mathcal{D})$ given by
\begin{equation}
C = w^xc(x) + w^d \sum_{i=1}^n c^i(d^i)
\label{eq:mdmp}
\end{equation}
\noindent such that
\begin{enumerate}
\item $x(0)=x^0$, $x(1)=x^g$
\item $\bigcup_{i \neq j} \left( \ o^i(d^i) \cap o^j(d^j) \ \right) = \{\emptyset\} \ \ \forall i,j \in \{1,\ldots,n\}$
\item $x(t) \cap \bigcup_{i=1}^n o^i(d^i) = \{\emptyset\}$, $\forall t \in [0,1]$ 
\end{enumerate}  
\label{def:mdmp}
\end{definition}
In~(\ref{eq:mdmp}), $c(x)$ is a function of the path length, $c^i(d^i)$ are positive displacement costs with $w^x$ and $w^d$ being the corresponding weights and $o^i(d^i)$ denotes the new obstacle locations. The first condition satisfies the endpoint constraints. The second condition ensures that the displaced obstacles are themselves collision free and the last condition guarantees the robot to not intersect with the displaced obstacles.

%% file: approach.tex
In this section we present our planner for minimum displacement motion planning that employs a two stage optimization approach, namely, (a) the planning phase and (b) the refinement phase. 
\subsection{(a) Planning Phase}
Let us consider a robot with a general dynamics model of the form
\begin{equation}
x_{k+1} = f(x_k,u_k) 
\label{eq:dynamic}
\end{equation}
\noindent where $x_{k+1}$, $x_k$ are the robot states at time $k+1$, $k$ respectively, and $u_k$ is the applied control at $k$. We define a function $V(x',o^i)$ to measure the robot-obstacle intersection or overlap, where $x' = g(x) $ denote the relevant robot state (at any time instant) of interest and $o^i$ is the obstacle state of interest. For example, if the robot and obstacle are spheres of radii $r^r$ and $r^o$, respectively, then $V$ is defined as
\begin{equation}
   V(x',o) = 
   \begin{cases}
     md \ &\text{if} \ \norm{x'-o}  \leq (r^r+r^o)\\
     0 \ &\text{otherwise}.
   \end{cases}
   \label{collcond}
\end{equation}
\noindent where 
\begin{equation}
md = \norm{x'-o} - (r^r+r^o)
\end{equation}
\noindent with $x'$, $o$ denoting the centers of the robot and obstacle spheres, respectively and $\norm{\cdot}$ denoting the $L_2$ norm. We note that if two spheres intersect/collide then $\norm{x'-o} \leq (r^r+r^o)$. Thus $V(x',o)$ is essentially the collision condition and measures the extent of intersection or overlap when the robot and the obstacle collide. It is easily observed that the magnitude of $md$, that is, $|md|$ is actually the minimum distance by which the obstacle sphere is to be moved along the line joining the robot and obstacle sphere centers so as to make this overlap equal to zero. For non spherical objects, $x'$ mostly comprises of the vertices and the center of the polyhedron and the $V(\cdot)$ is the collision condition between the two polyhedrons. However, in this paper we use minimum volume spheres as the underlying bounding volumes to tightly enclose a polyhedron as they provide an efficient approximation.

At each time instant $k$, the robot plans for $L$ look-ahead steps minimizing an objective function
\begin{equation}
J_k = \sum\limits_{l=0}^{L-1} c_l(x_{k+l},\mathcal{O}) + c_L(x_{k+L})
\end{equation}
\noindent with $c_l$ being the cost for each look-ahead step and $c_L$ the terminal cost. As a result, we have the following immediate and terminal costs
\begin{eqnarray}
c_l(x_{k+l},\mathcal{O}) = \norm{x_{k+l}}^2_{M_x} + \sum_{i=1}^n \norm{V^i}^2_{M_i} \\
 c_L(x_{k+L}) = \norm{x_{k+L} - x^g}^2_{M_g} 
 \end{eqnarray}
\noindent where $\norm{x}_S = \sqrt{x^TSx}$ is the Mahalanobis norm, $M_u, M_g, M_i$ are weight matrices. The optimization problem can now be formally stated as 
\begin{mini!}|s|[1]
{ }{ \sum\limits_{l=0}^{L-1}\left[ \norm{x_{k+l}}^2_{M_x} + \sum_{i=1}^{n} \norm{V^i}^2_{M_i} \right]  + \norm{x_{k+L} - x^g}^2_{M_g} }{\label{eq:cost_fn}}{}
  \addConstraint{x_0}{=x^0}
  \addConstraint{x_{k+l}}{=f(x_{k+l-1},u_{k+l-1})}
  \addConstraint{x_T}{=x^g}
    \addConstraint{u_{k+l}}{\in U}{\label{eq:controls}}
      \end{mini!}
\noindent where~(\ref{eq:controls}) constraints the control actions to lie within the feasible set of control inputs. The term $\norm{V^i}^2_{M_i} = M_i (md^i)^2$ penalizes the robot-obstacle intersections and the optimization returns the robot trajectory $x_0,...,x_T$ with minimum obstacle overlaps $md^i$ at each time instant. As discerned before, the magnitude of $md^i$ is actually the minimum displacement of the obstacle $o^i$ required to make the robot-obstacle overlap equal to zero. However, each obstacle $o^i$ is not overlapped by a single robot state $x_k$ but $l$ different discretized states since the robot trajectory is continuous. Moreover, a robot and the obstacle may be represented by $m^1$ and $m^2$ different bounding volume spheres, respectively. As a result, for each obstacle $o^i$, we have $md^{i^1},\ldots,md^{i^p}$, $1\leq p \leq l\times m^1 \times m^2$ different overlaps with the robot. The maximum among them is the required distance for zero overlap.
The planning phase is summarized below:\\\\
\textbf{Planning Phase:}\\
1. Input: $x^0$, $x^g$, $\mathcal{O}$,$y^1=0,\ldots,y^n=0.$\\
2. Compute $x_0,...,x_T$ such that $J_k$ is minimized at each $k\in[0,T]$.\\
3. Simulate $x_0,...,x_T$ to compute $ \{md^{i^j}\},\ 1\leq j\leq p $ for each $o^i$. \\
4. Compute $y^i = \max(\{md^{i^j}\})$, to obtain the distance for each obstacle $o^i$.\\\\
Though we have computed the minimum distances to be moved, we have not yet ensured the condition 2., 3. in Definition~\ref{def:mdmp}, that is, the displaced obstacles being themselves collision free and the robot not colliding with the displaced obstacles while executing the computed trajectory. The refinement phase of the planner attends to these concerns.
\subsection{(b) Refinement Phase}
Displacing the obstacles by $y^i$ does not ensure the obstacles being themselves collision free since it may introduce new overlaps with the obstacles. Similarly, in some cases moving the obstacle by $y^i$ can lead to new intersections with the remaining robot trajectory. To tackle this, we iteratively increase the displacement magnitudes.

The refinement phase is essentially a displacement sampling process. We note here that the $y^i$ computed is the minimum distance the obstacle sphere is to be moved along the line joining the robot and obstacle spheres. We note here that though a given obstacle may be represented using multiple spheres, $y^i$ computed represents the maximum displacement from among all these spheres. The obstacle center is thus displaced by $y^i$ along the maximum displacement direction. Each displaced obstacle is checked for collision with the robot and the other obstacles. If no displaced obstacles collide with the robot and other obstacles then the obtained $y^i$ are the required minimum displacements. However, most often due to the nature of the environment and the continuous robot trajectory, some of the displaced obstacles introduce new overlaps, that is, collide either with the robot or other obstacles or both. A displacement $y^i = y^i + \Delta$ is computed and obstacle locations at a distance equal to the newly computed $y^i$ from the initial obstacle location $o^i$ are sampled and checked for collision. The displacements are incremented by $\Delta$ until no collisions are found. The refinement process is as follows:\\\\
\textbf{Refinement Phase:}\\
1. Input: $x_0,...,x_T$, $\mathcal{O}$,$y^1,\ldots,y^n$, $\Delta$, $d^1=0,\ldots,d^n=0$.\\
2. While \textit{true}\\
3. For each $o^i$ such that $md^{i^j} \neq 0$\\
4. Find $y^i$.\\
5. \hspace{0.3cm}Run \textit{Sample-Displacement}.\\
6. \hspace{0.3cm}For $sample$\\
7. \hspace{0.6cm}If $\bigcup_{i \neq j} \left( \ o^i(y^i) \cap o^j(y^j) \ \right) = \{\emptyset\} \ \ \forall i,j \in \{1,\ldots,n\}$ and $x_t \  \cap \ \bigcup_{i=1}^n o^i(y^i) = \{\emptyset\} \ \forall t \in \{1,\ldots,T\} $.\\
8. \hspace{0.9cm}$d^i = y^i $.\\
9. \hspace{0.6cm}Else\\
10. \hspace{0.9cm}$y^i = y^i + \Delta$.\\\\
The \textit{Sample-Displacement} subroutine samples obstacle locations at a distance of $y^i$ from the initial obstacle location $o^i$. For a 2D scenario, the samples lie on a circle with center being the initial obstacle location before displacement, that is, $o^i$, and radius equal to the newly computed $y^i$. In a 3D scenario, the circle is replaced by a sphere and samples are points on the surface of the sphere. If the obstacle can only be rotated, then a discrete set of rotations are sampled. The subroutine runs as follows:\\\\
\textbf{Sample-Displacement:}\\
1. Input: $\mathcal{O}$, $y^i$.\\
2. For $N=1,2,\ldots$\\
3. \hspace{0.3cm}If translation\\
4. \hspace{0.6cm}Sample $o^i(y^i)$ such that $\norm{o^i(y^i) - o^i} = y^i$.\\
5. \hspace{0.3cm}If rotation   \\
6.  \hspace{0.6cm}Sample $o^i(y^i)$, such that $0\leq y^i < 2\pi$.\\
7. Return \textit{sample} $o^i(y^i)$.
\subsection{Optimality}
The cost given in~(\ref{eq:mdmp}) is a combination of the sum of the costs in the planning phase and the incremental displacements computed in the refinement phase. Thus,
\begin{equation}
C = \sum_{k=0}^T \left(J_k + \sum_{i=1}^n (d^i-y^i)\right)
\end{equation}
To prove any sort of optimality of our two stage planner, it should exhibit the optimal substructure property. A problem has an optimal substructure only if an optimal solution to the problem can be obtained from optimal solutions to its subproblems~\cite{cormen2009book}. We assume that we have a mechanism by which we obtain the optimal values for the weights $M_u, M_g, M_i$.  Let $M_u^\star, M_g^\star, M_i^\star$ give the optimal cost $J^\star = \sum_{k=0}^T J_k$. Let the distances returned by planning phase be $\{y^{i^{\star}}\}$. In the ideal scenario, that is, $\bigcap_{i=1}^n o^i(y^{i^{\star}}) = \{\emptyset\}$ and $\left(\bigcap_{t=1}^T x_t \right) \cap \left(\bigcap_{i=1}^n o^i(y^{i^{\star}})\right) = \{\emptyset\}$, and no refinement phase is required. Thus, $d^i = y^{i^{\star}}$. $J^\star$ is optimal because if there were another $J^+ \leq J^\star$, then we would use $J^+$ since the optimal weights $M_u^+, M_g^+, M_i^+$ are known to us.

If the refinement phase is to be run, that is, $d^i \neq y^{i^{\star}}$, then we find $d^i$ using incremental sampling. This $d^i$ can be made asymptotically closer to the actual value by choosing a sufficient increment $\Delta >0$. Thus, for any $\epsilon >0 $ that forms an allowable limit on the suboptimality, $d^i \rightarrow d^{i^{\star}} + \epsilon $ with probability 1. Note that this is a direct consequence of our \textit{Sample-Displacement} subroutine. The optimal cost is thus obtained as $C^\star = J^\star + \sum_{i=1}^n (d^i-y^{i^{\star}})$. 

Let us consider any other solution $J'$ for the planning phase with $M'_u, M'_g, M'_i$ being the weights. Based on our assumption, any weights other than $M_u^\star, M_g^\star, M_i^\star$ gives $J'>= J^\star$ and $\sum y^{i'} >= \sum y^{i^{\star}}$. Now since the refinement phase proceeds in an incremental manner, we have $\sum (d^i -y^{i'}) >= \sum (d^i - y^{i^{\star}})$. Therefore we have, $C' = J' + \sum_{i=1}^n (d^i-y^{i'}) >= C^\star$, giving the desired result.

%% file: examples.tex
The nonlinear model predictive control (NMPC) based optimization in~(\ref{eq:cost_fn}) is realized using the solver FORCESPRO~\cite{FORCESPro,FORCESNlp} which generates optimized NMPC code. Note that the term $V^i$ in the objective function~(\ref{eq:cost_fn}a) is a conditional penalty term as defined in~(\ref{collcond}). Since the objective function has to be continuously differentiable (smooth), the switch between the penalty value and zero will have to be performed in a smooth manner. To this end, we employ a built-in smooth approximation function provided by FORCESPRO that computes the minimum between two quantities\footnote{\url{https://forces.embotech.com/Documentation/modelling/index.html?\#smooth-minimum}}. The optimal weights $M_x$, $M_i$, $M_g$ vary with each problem and the tuning is currently achieved by trial and error. For all the experiments, we use a look-ahead $L=21$ and an incremental displacement $\Delta = 0.01$. The offline code generation using FORCESPRO (planning phase) took around 16 seconds for the \textit{IAS domain} with 35 obstacles. The performance is evaluated on an Intel{\small\textregistered} Core i7-10510U CPU$@$1.80GHz$\times$8 with 16GB RAM under Ubuntu 18.04 LTS.

\subsection{Theoretical Examples}
The L-shaped robot example introduced in Fig.~\ref{fig:L} has the following non-linear dynamics model
\begin{equation}
\begin{split}
\dot{x}_x & = ucos(\theta) - vsin(\theta)\\
\dot{x}_y & = usin(\theta) + vcos(\theta)\\
\dot{\theta} & = \omega\\
\end{split}
\label{L-model}
\end{equation}
\noindent where $x = (x_x,x_y,\theta)$ is the robot state (location and orientation) with $u,v$ being the linear velocities and $w$ being the angular velocity. Fig.~\ref{fig:L2} shows the shortest path trajectory with the planning phase displacement $\sum y^i = 13.08$. The refinement phase gave the final displacement of $\sum d^i = 18.52$ which is observed in Fig.~\ref{fig:L2}. On the other hand, Fig.~\ref{fig:L3}  in which the displacements are heavily penalized gave $\sum y^i = 7.53$ with a final total displacement of $\sum d^i = 12.12$. Fig.~\ref{fig:L4} shows an instance of the planning phase (line 3) where the computed trajectory is simulated to obtain the set of $\{md^{i^j}\}$, that is, the set of robot-obstacle overlaps. 
\begin{figure}[t]
  \subfloat[]{\includegraphics[trim=52 110 38 100,clip,scale=0.5]{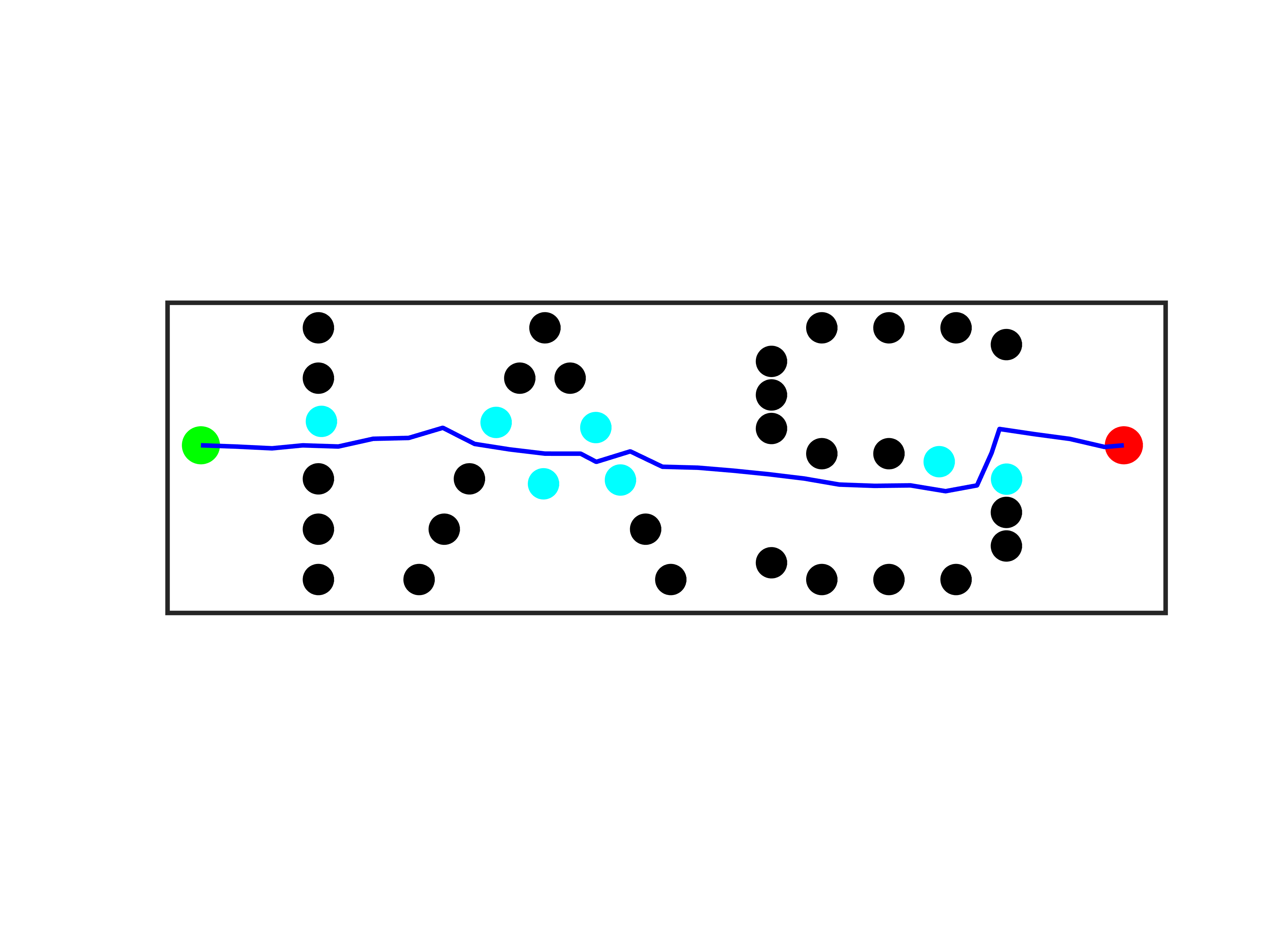}\label{fig:C1}}\hfill
  \subfloat[]{\includegraphics[trim=52 110 38 100,clip,scale=0.5]{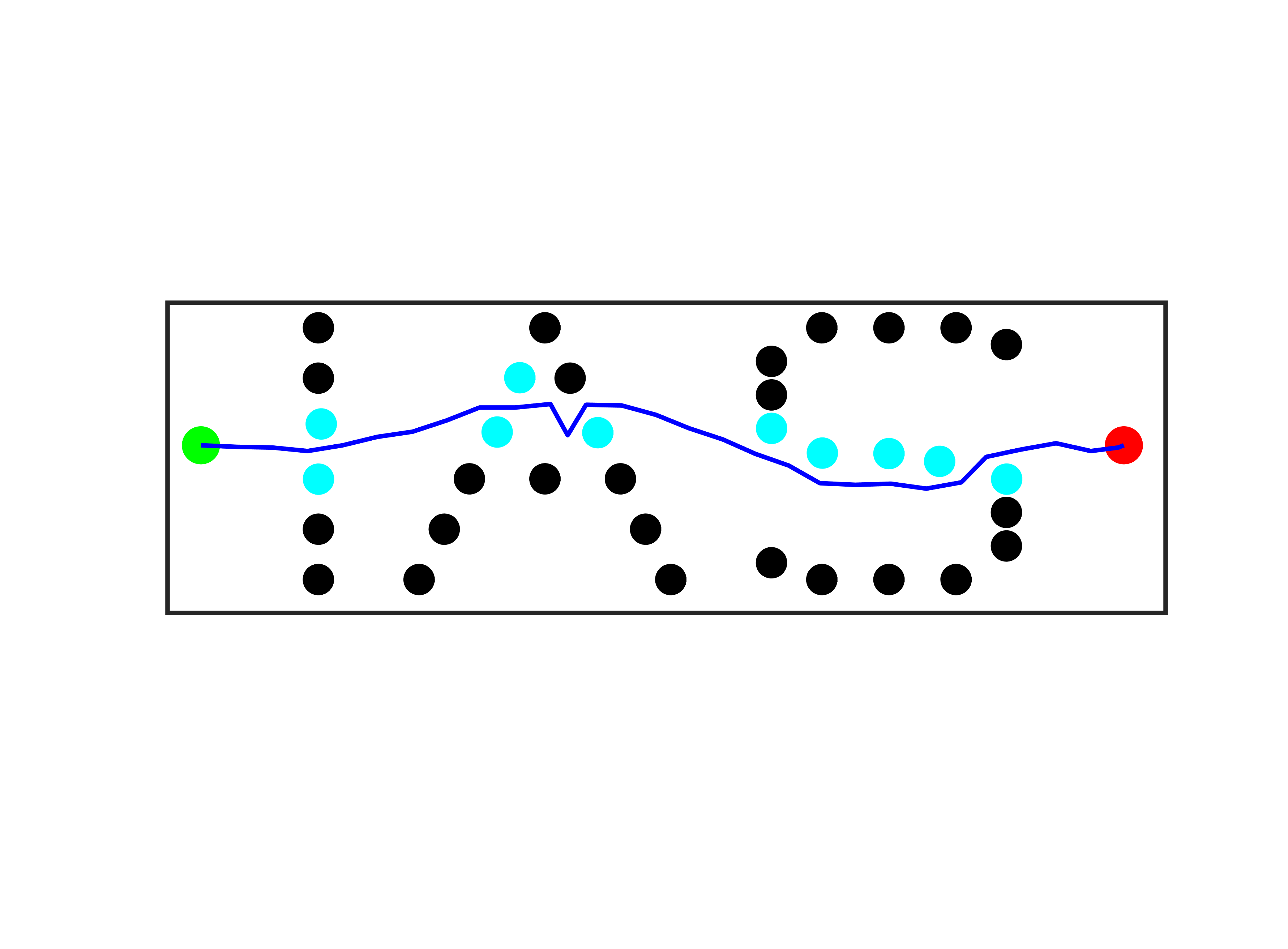}\label{fig:C2}}\hfill
    \caption{A robot moving from left to right in the \textit{IAS domain}. (a) A 7-obstacle solution. (b) A 10-obstacle solution with smaller displacement.}
  \label{fig:C}
\end{figure}

Fig.~\ref{fig:C} shows two instances of the IAS domain where 7 (Fig.~\ref{fig:C1}) and 10 (Fig.~\ref{fig:C2}) obstacles are displaced while the robot reaches its goal. Fig.~\ref{fig:C1} gave a planning phase displacement of $\sum y^i = 1.19$ and refinement phase displacement of $\sum d^i =1.32$. However, Fig.~\ref{fig:C2} gave $\sum y^i = 1.01$ and a final total displacement of $d^i = 1.05$. This example demonstrates that minimum displacement of obstacles does not always imply minimum number of obstacles being displaced. We use the following non-linear dynamics for the robot
\begin{equation}
x_{k+1}= \begin{bmatrix}
x_x  \\ 
x_y  \\
\theta \\
\end{bmatrix}_{k+1} = 
\begin{bmatrix}
x_x  \\ 
x_y  \\
\theta \\
\end{bmatrix}_{k} + 
\begin{bmatrix}
D cos(\theta_k + T/2) + C cos(\theta + (T+\pi)/2)  \\ 
D sin(\theta_k + T/2) + C sin(\theta + (T+\pi)/2)  \\
T  \\
\end{bmatrix}
\label{C-model}
\end{equation}
\noindent where $x = (x_x,x_y,\theta)$ is the robot state and $D$, $C$, $T$ are the down-range, cross-range and turn control components, respectively. 

\begin{figure}[t]
  \subfloat[]{\includegraphics[trim=52 80 38 70,clip,scale=0.5]{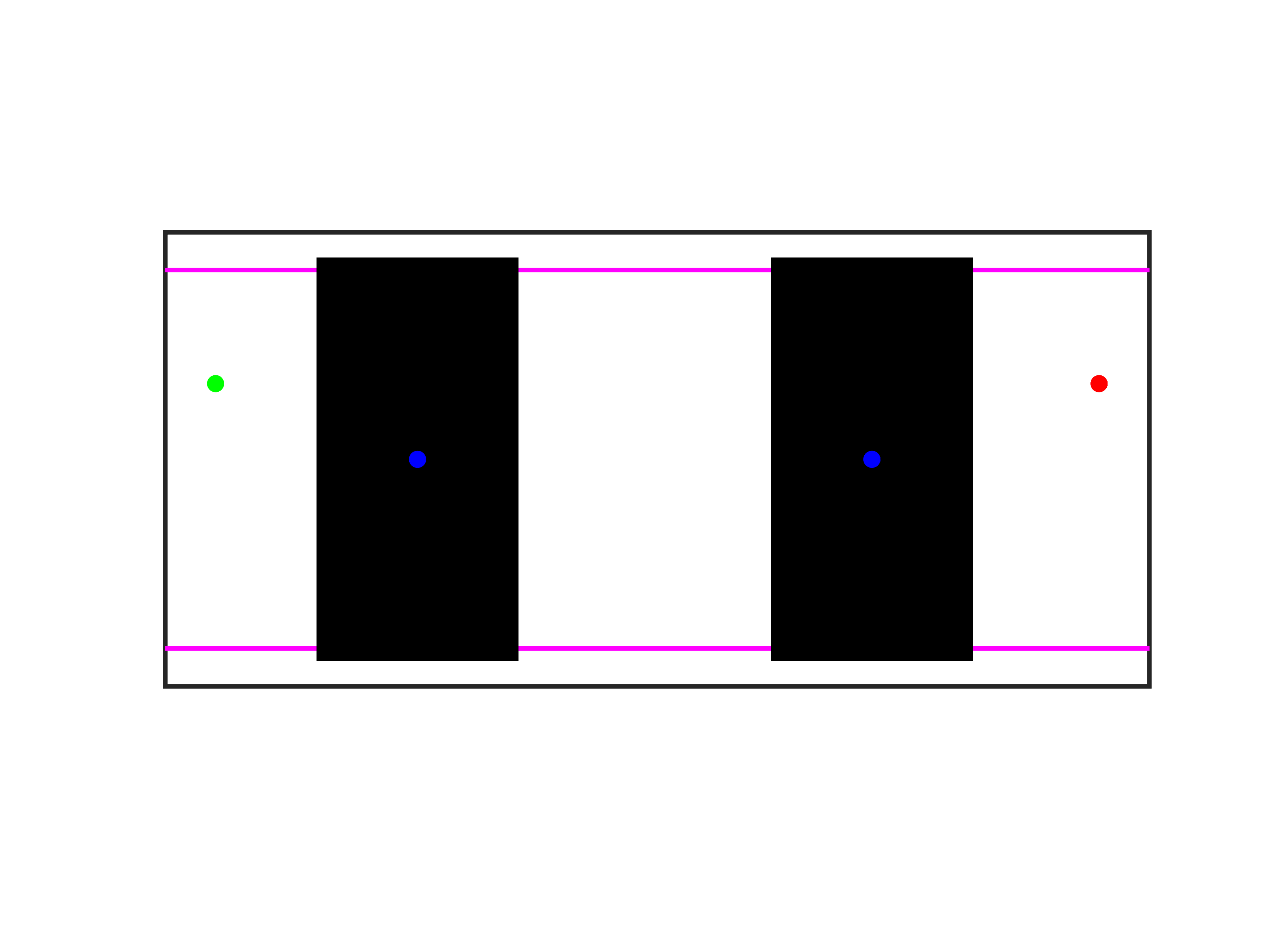}\label{fig:rot1}}\hfill
   \subfloat[]{\includegraphics[trim=52 80 38 70,clip,scale=0.5]{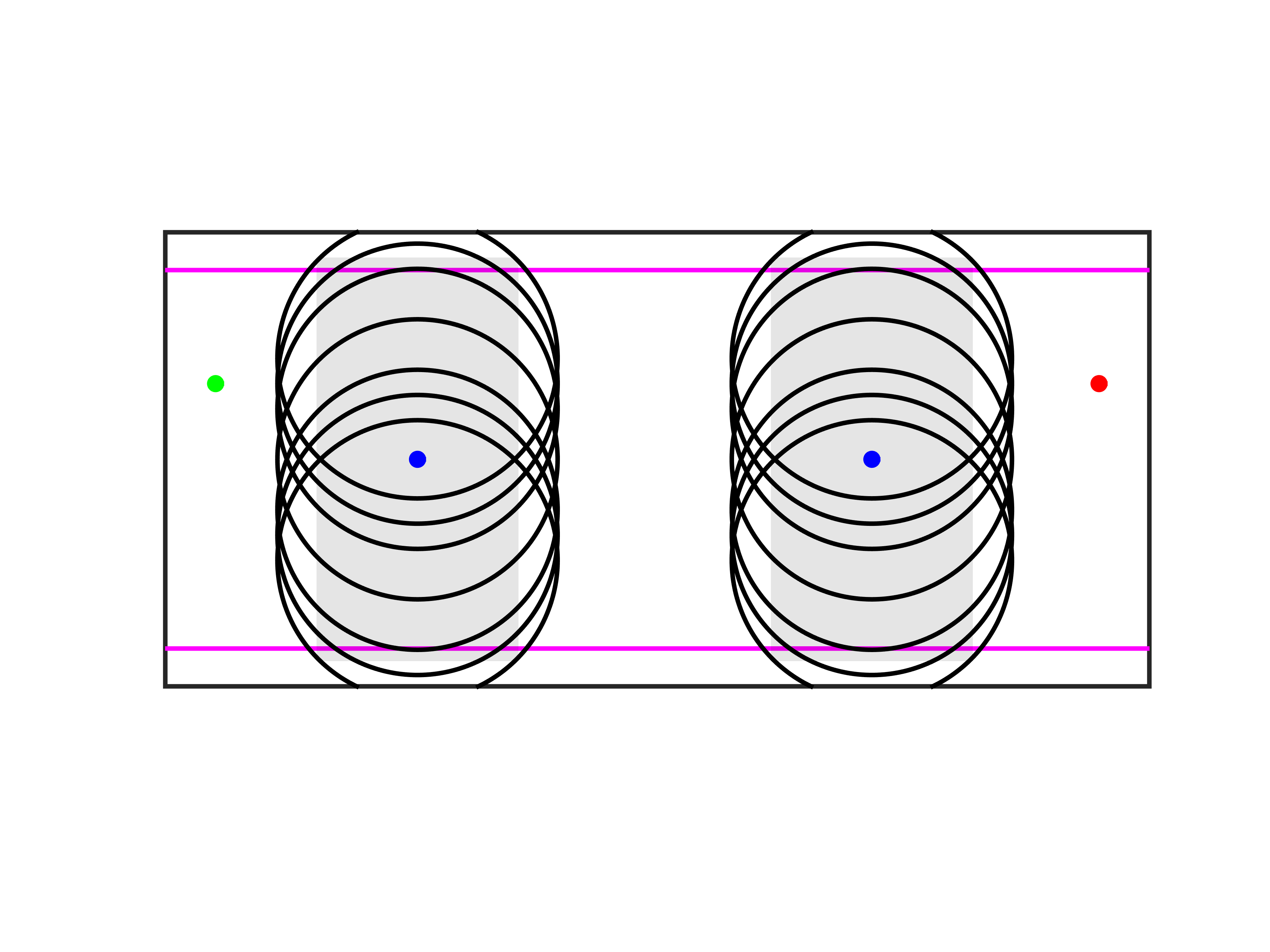}\label{fig:rot2}}\hfill
  \subfloat[]{\includegraphics[trim=52 80 38 70,clip,scale=0.5]{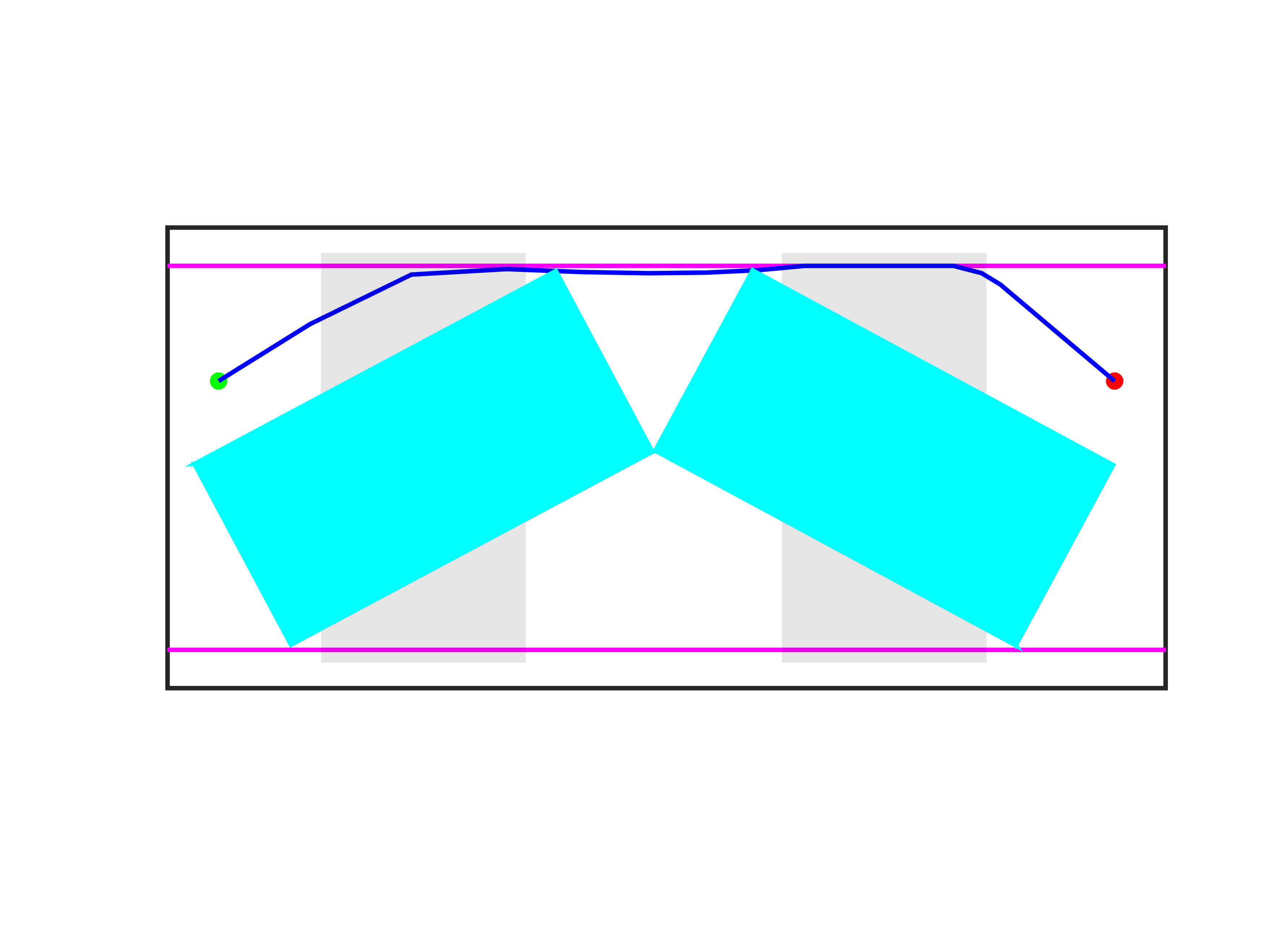}\label{fig:rot3}}\hfill
  \subfloat[]{\includegraphics[trim=52 80 38 70,clip,scale=0.5]{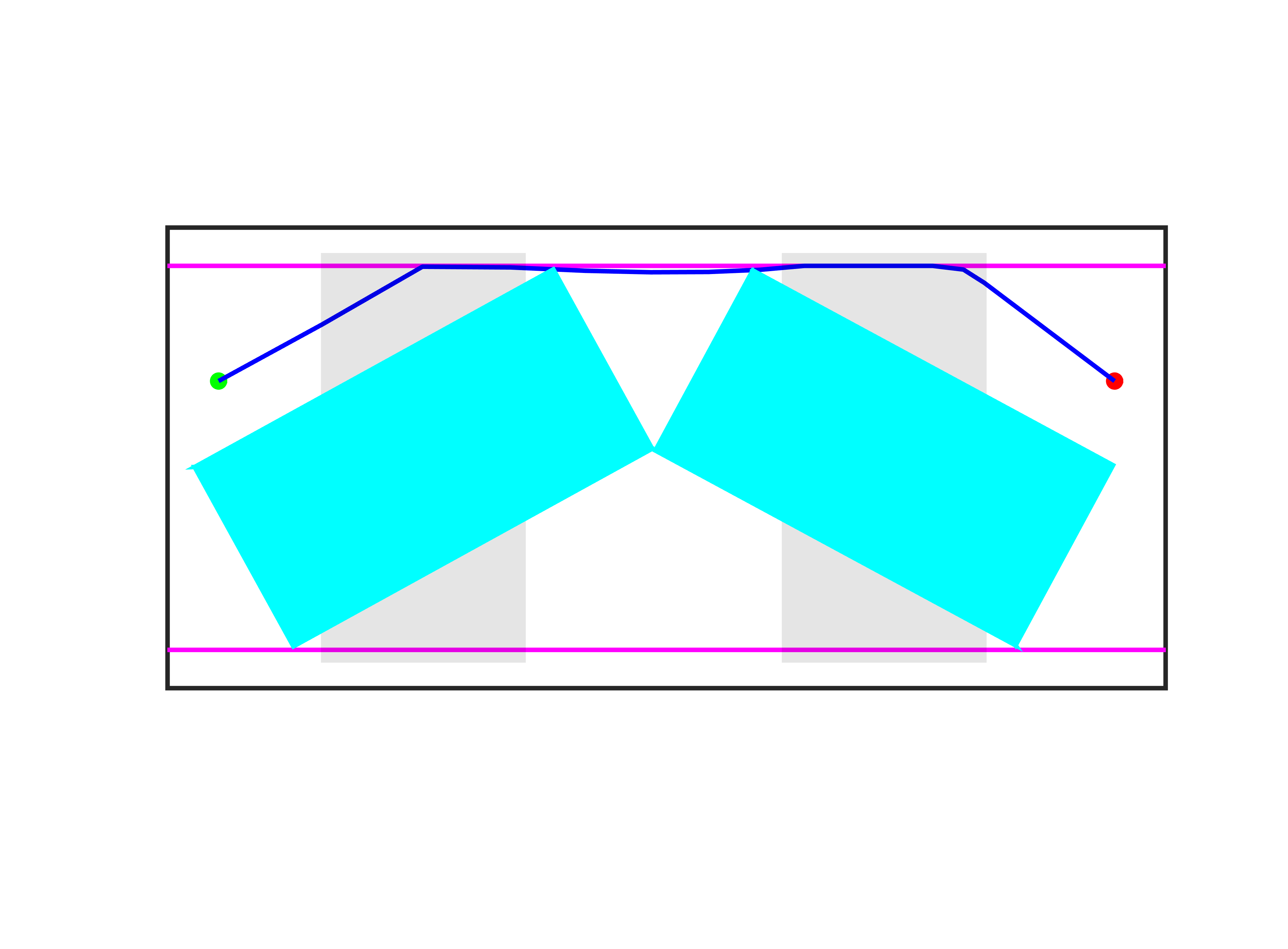}\label{fig:rot4}}\hfill
     \caption{(a) A point robot needs to move from left to right by moving the two blocks which can only be rotated about its centers (blue blobs). (b) Bounding volume spheres used to enclose the obstacles are shown. (c) Solution obtained using our approach. (d) Assumption of bounding volume spheres is relaxed and the solution obtained by minimizing robot-rectangle overlap is shown.}
  \label{fig:rot}
\end{figure}

Fig.~\ref{fig:rot} exhibits an example of pure rotation. A point robot has to reach its goal by displacing two blocks that can only be rotated about their centers. The point robot follows the dynamics given in~(\ref{C-model}) and its $y-$axis component $x_y$ is constrained to lie between the magenta lines. Fig.~\ref{fig:rot2} shows the bounding volume spheres used for both the blocks and Fig.~\ref{fig:rot3} displays the corresponding solution. The total rotation computed is $\sum_{i=1}^2 d^i =\sum_{i=1}^2 y^i = 1.01$. We now relax the assumption of bounding volume spheres by using the collision condition between a rectangle and a point in~(\ref{collcond}) and compute the solution by penalizing the robot whenever it is inside the rectangle. The resulting trajectory is given in Fig.~\ref{fig:rot4} with a total displacement of $0.98$. Thus, our use of bounding volume spheres for enclosing obstacles is a reasonable assumption.  
\subsection{Applications}
\begin{figure}[h]
\centering
 \subfloat[]{\includegraphics[width=0.32\textwidth]{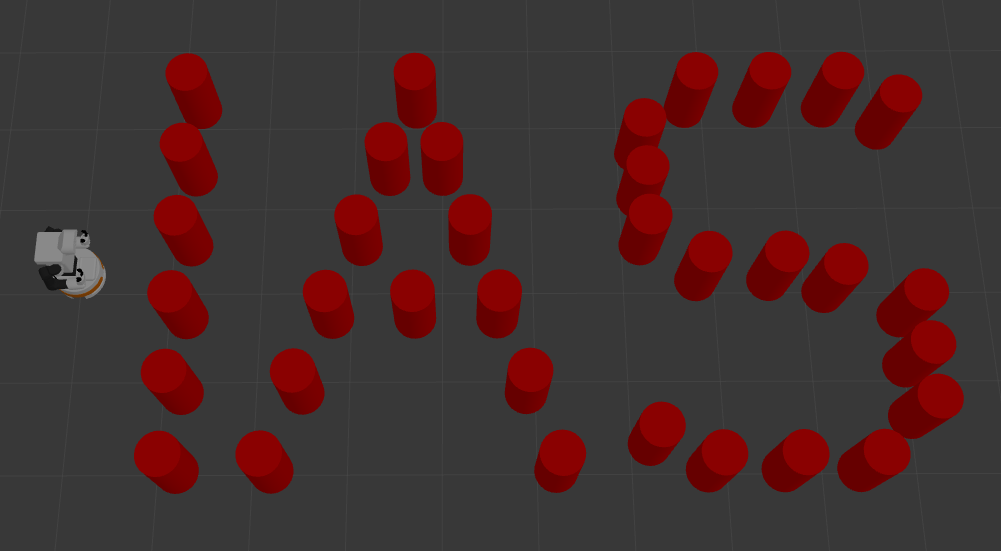}}\hfill
  \subfloat[]{\includegraphics[width=0.32\textwidth]{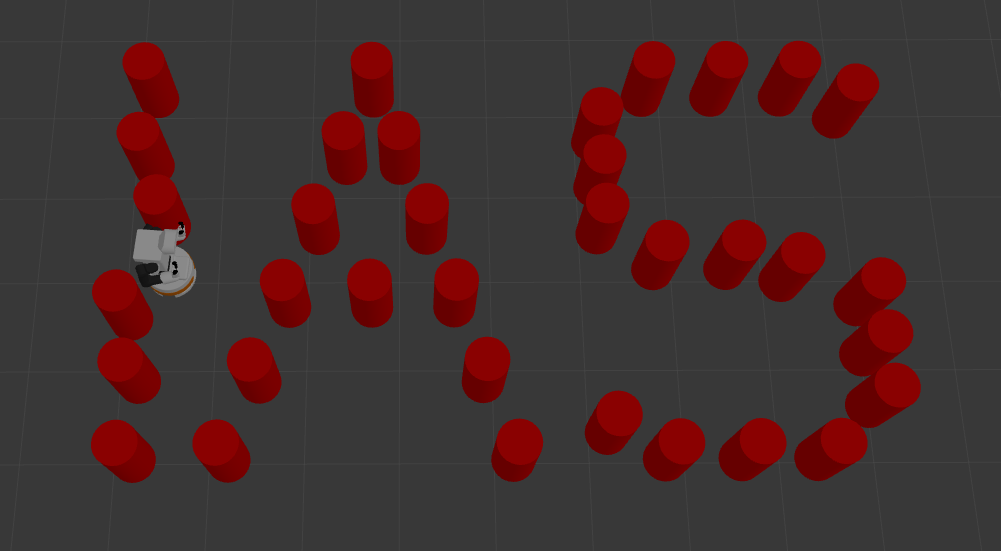}}   \hfill 
  \subfloat[]{\includegraphics[width=0.32\textwidth]{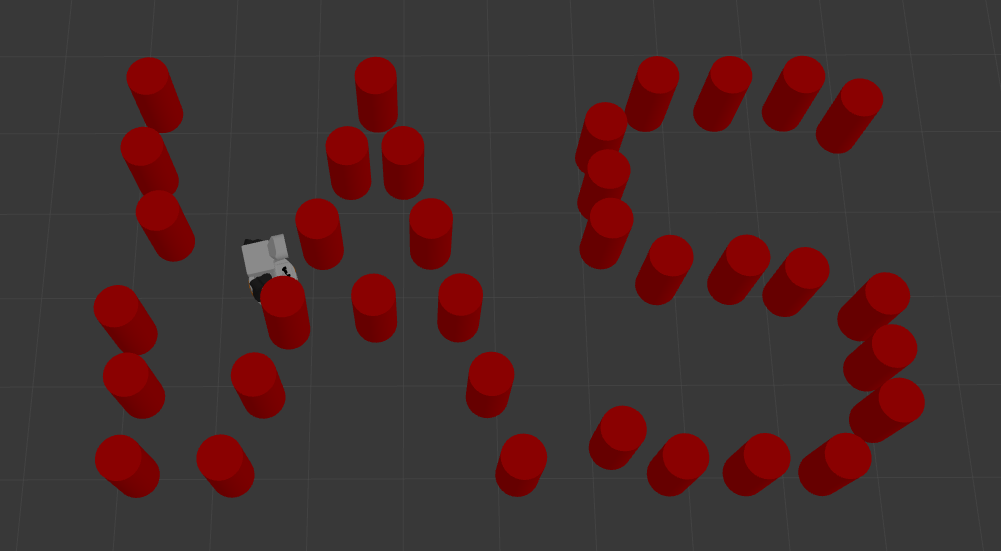}}\hfill
   
 \subfloat[]{\includegraphics[width=0.32\textwidth]{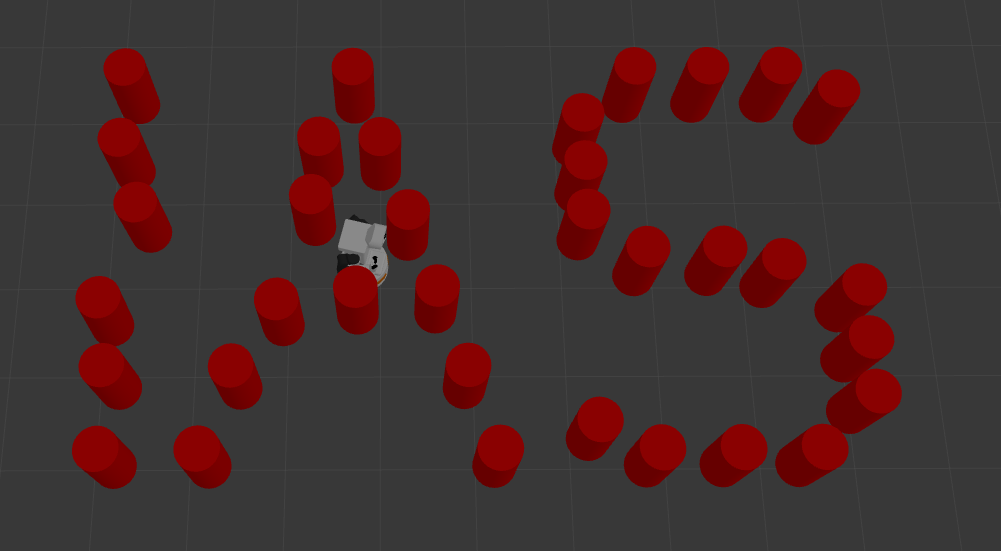}}\hfill
   \subfloat[]{\includegraphics[width=0.32\textwidth]{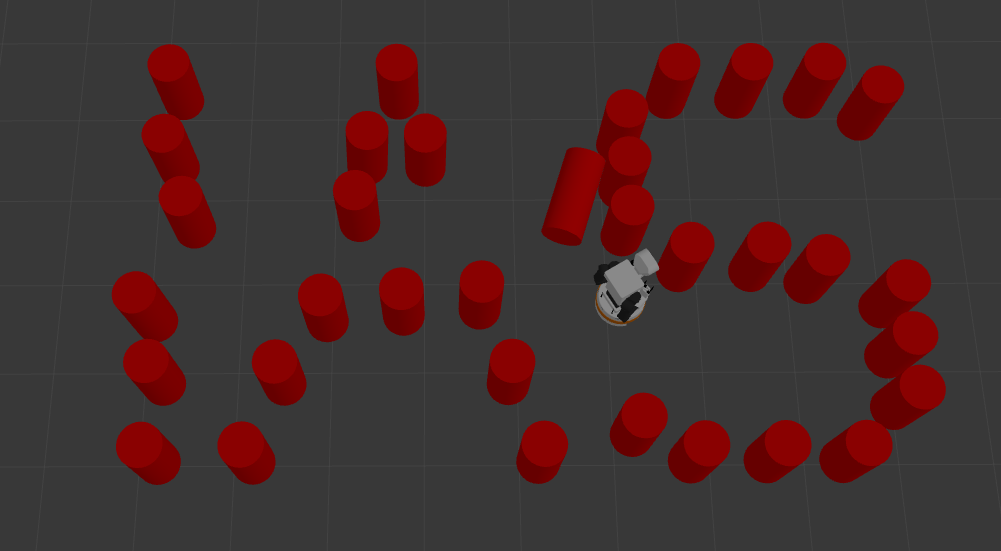}}\hfill
  \subfloat[]{\includegraphics[width=0.32\textwidth]{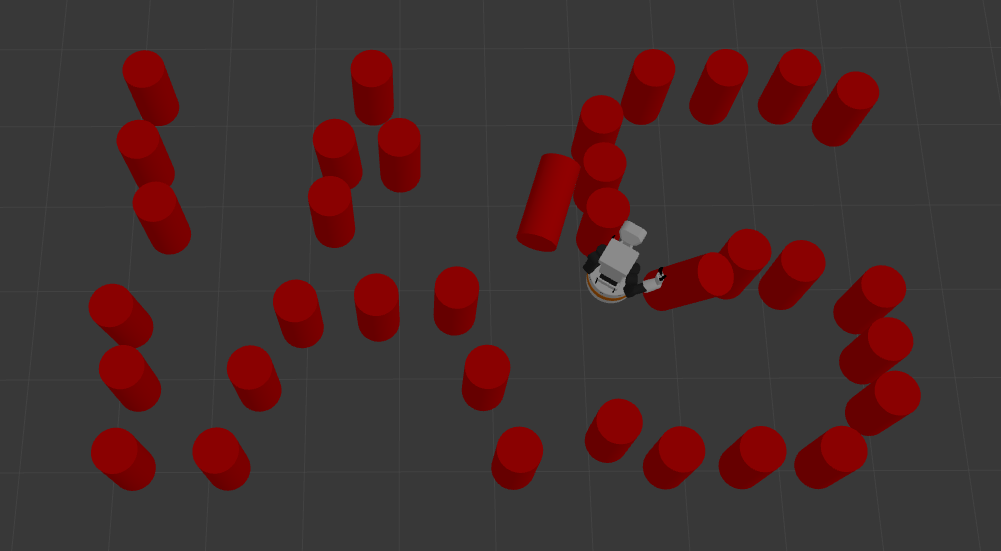}} \hfill   

  \subfloat[]{\includegraphics[width=0.32\textwidth]{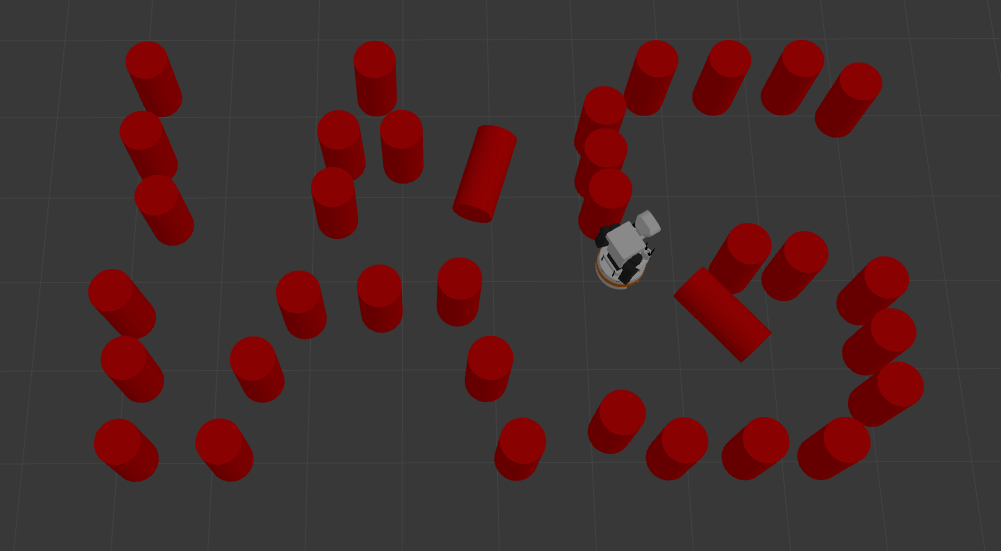}} \hfill
 \subfloat[]{\includegraphics[width=0.32\textwidth]{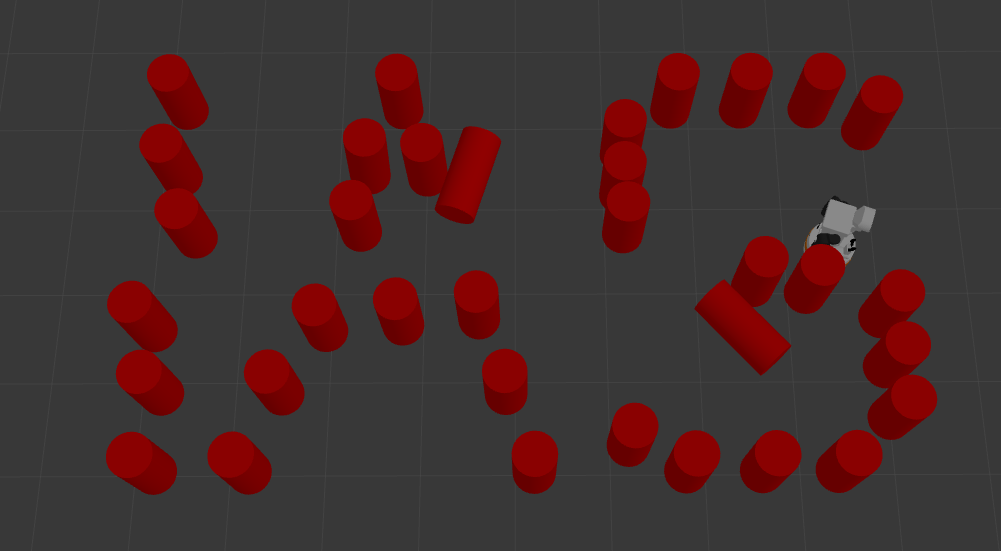}}\hfill
  \subfloat[]{\includegraphics[width=0.32\textwidth]{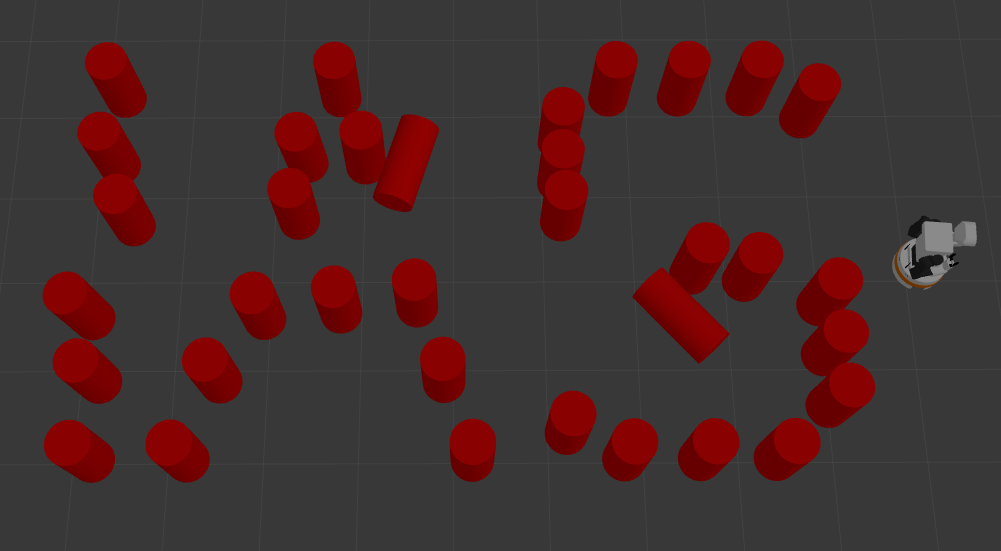}}\hfill
        \caption{Different instances of a robot navigating the \textit{IAS domain} by moving obstacles to reach its goal.}
  \label{fig:snapshot_multi}
\end{figure}

\begin{figure}[t]
    \centering
    \begin{tabular}{cc}
    \adjustbox{valign=b}{\subfloat[]{%
          \includegraphics[scale=0.16]{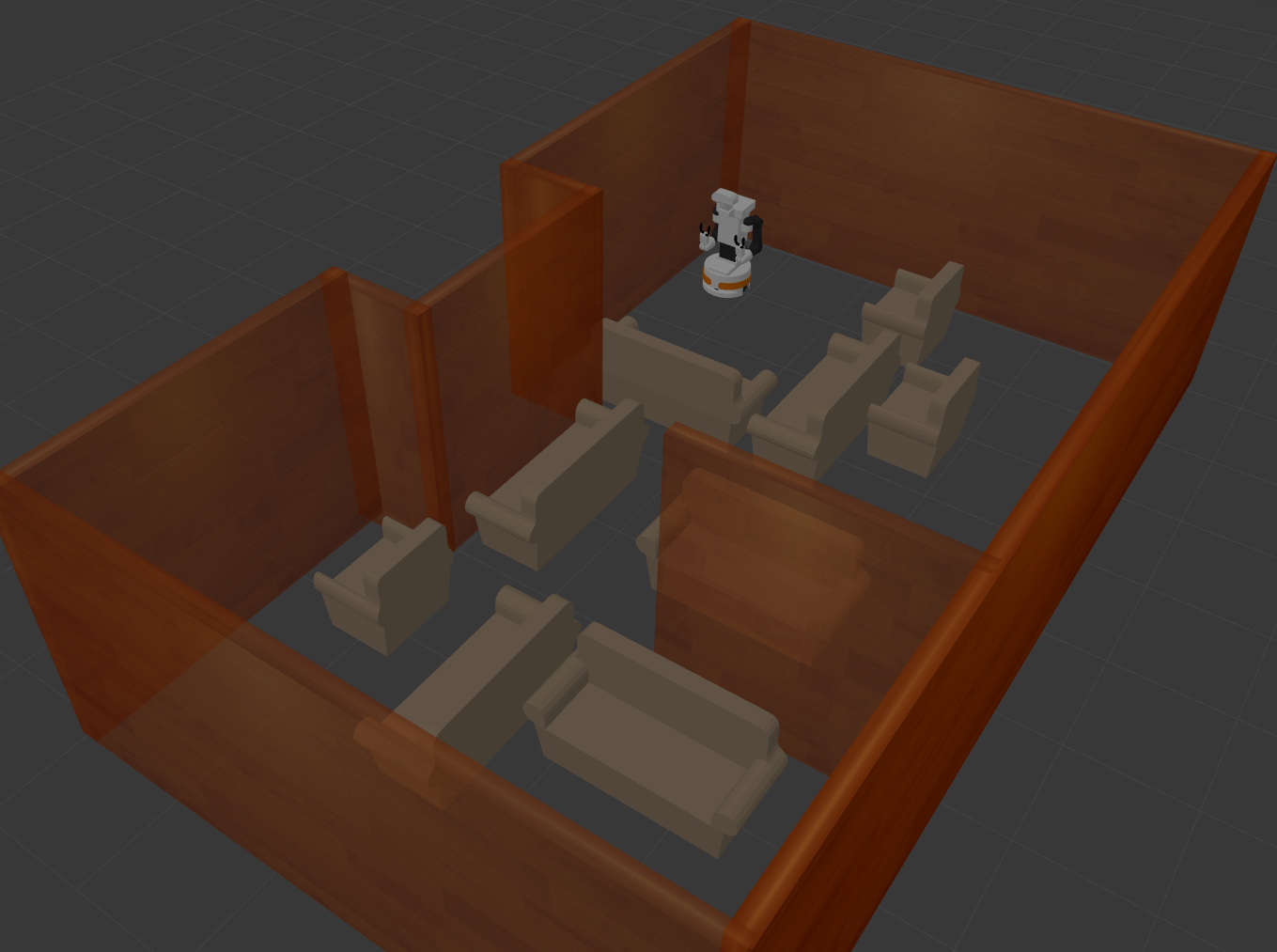}\label{fig:world1}}}
    &      
    \adjustbox{valign=b}{\begin{tabular}{@{}c@{}}
    \subfloat[]{%
          \includegraphics[trim=52 60 38 55,clip,scale=0.34]{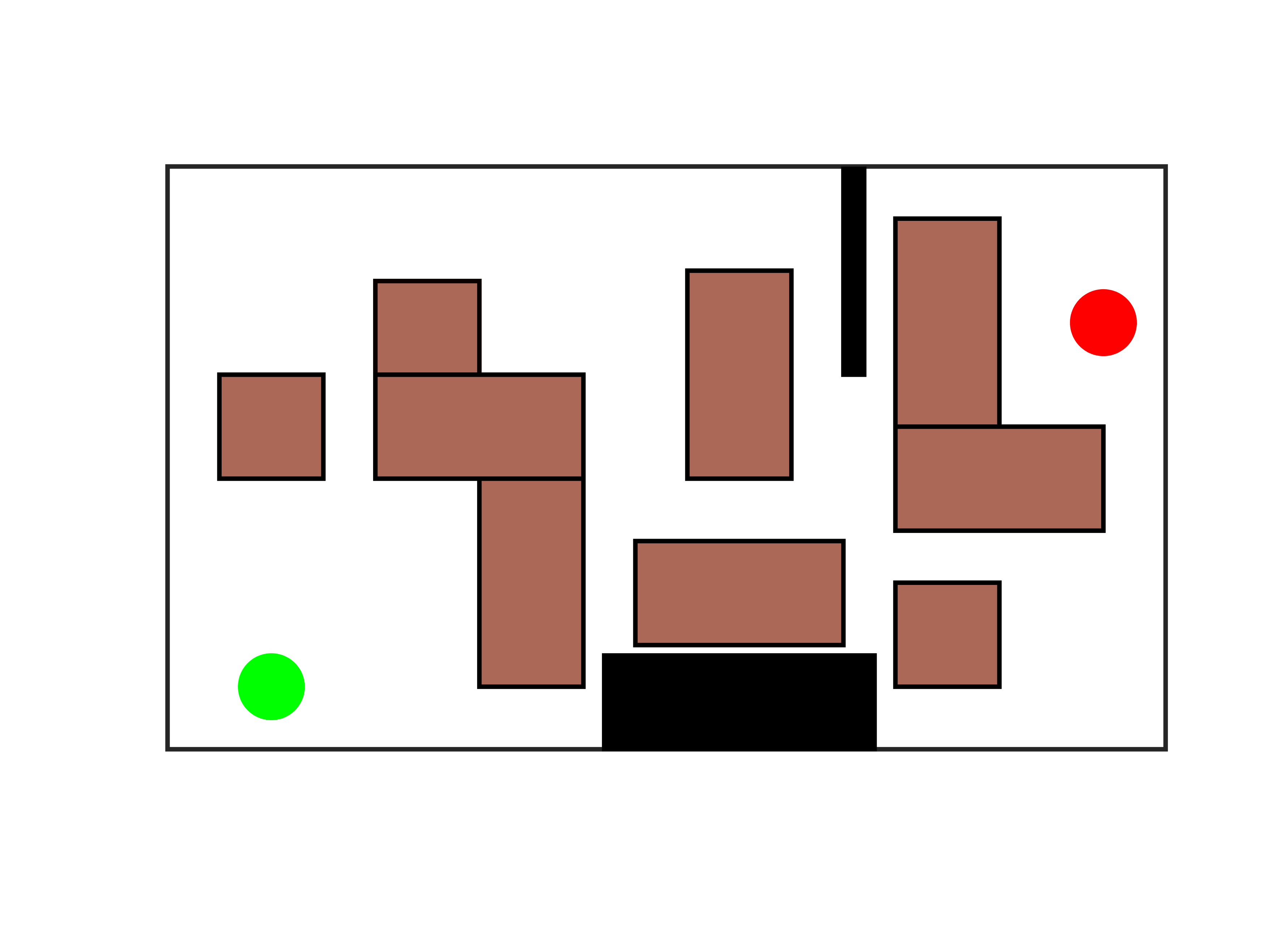}\label{fig:world2}} \\
    \subfloat[]{%
          \includegraphics[trim=52 60 38 55,clip,scale=0.34]{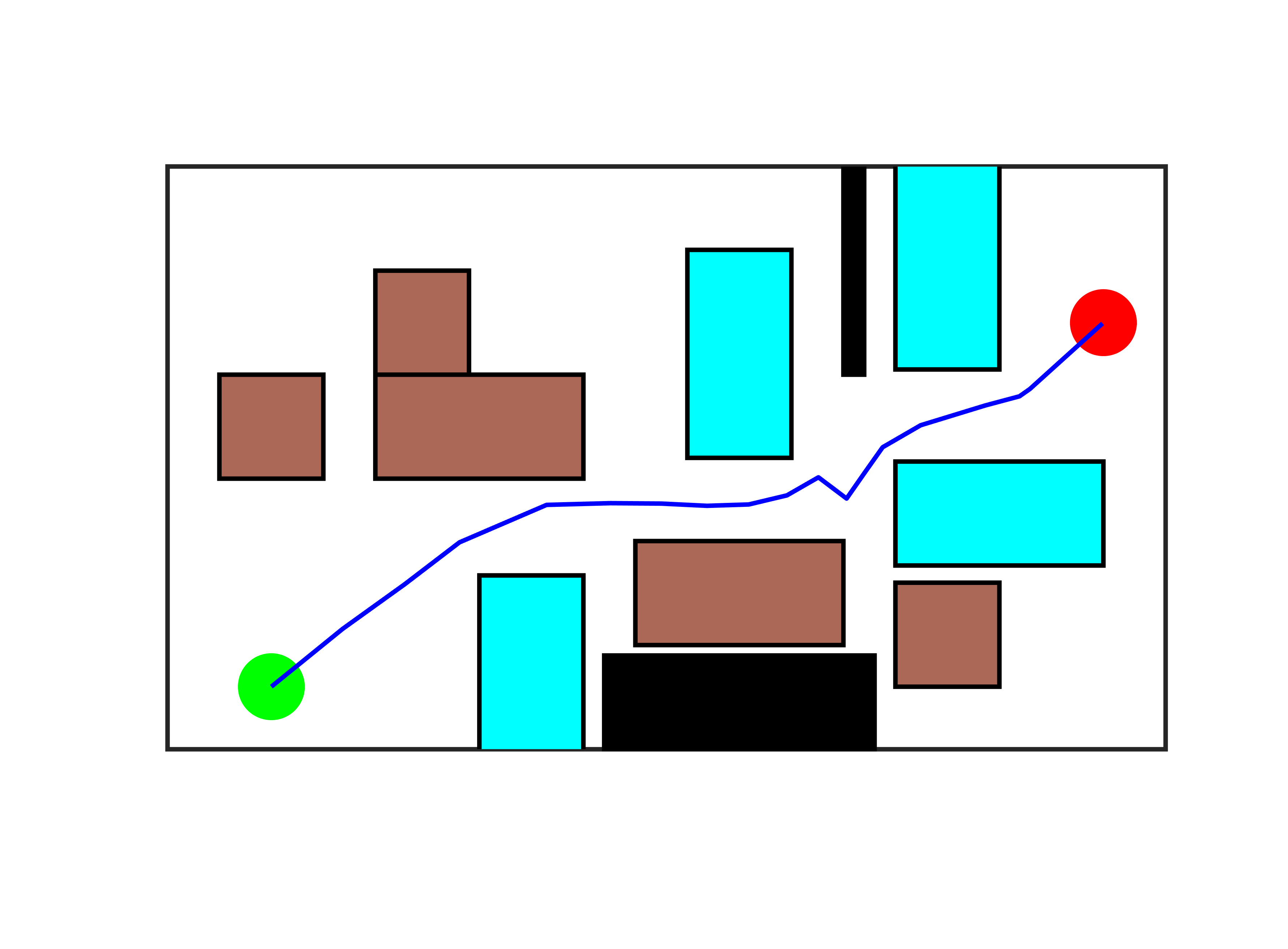}\label{fig:world3}}
    \end{tabular}}
    \end{tabular}
    \caption{(a) The \textit{sofa domain} where the robot has to reach the goal by push around the sofas. (b) Map of the domain with green denoting the starting location of the robot and red being the goal location. (c) 4-obstacle solution with the displaced sofas shown in cyan. }
  \label{fig:world}
  \end{figure}

There exists several applications wherein a robot need to move through a cluttered space by rearranging or repositioning objects along its way-- displacing chairs or other objects of interest. In Fig.~\ref{fig:snapshot_multi} different instances of a robot navigation through the \textit{IAS domain} among movable obstacles can be seen. The trajectory and the obstacle displacements are computed offline as described in Section~\ref{sec:approach}. As the robot moves along the computed trajectory the obstacles are moved aside with the help of its arms. In the experiment reported in Fig.~\ref{fig:snapshot_multi}, the arm motions to displace the obstacles are teleoperated by a human. 
Fig.~\ref{fig:world} shows the \textit{sofa domain} wherein the robot has to reach its goal by pushing around the sofas. The resulting 4-sofa solution with minimum displacement is shown in Fig.~\ref{fig:world3}.

%% file: related.tex
 Minimum displacement motion planning problem is not an entirely new class of problems and bears resemblance several existing works in the literature. If all the displacements equal to zero then we have an instance of the classical motion planning or the Piano Mover's problem~\cite{reif1979CS}. By tuning the weights $w^x$ and $w^d$ to heavily penalize longer paths, the problem can be reduced to finding the shortest path (see Fig.~\ref{fig:L}). Our work is closely related to the minimum constraint displacement (MCD)~\cite{hauser2013RSS} and minimum constraint removal (MCR)~\cite{hauser2014IJRR} motion planning problems. Both MCD and MCR sample robot configurations and incrementally grows the roadmap (robot configurations connected by straight lines). We do not sample robot configurations and plan a path employing the actual robot dynamics to minimize the robot-obstacle overlaps. MCD also sample obstacle displacements and the iterations (extension of roadmap and displacement sampling) continue until the best cost solution is obtained. MCD does not consider obstacle interactions and the displaced obstacles are allowed to overlap with each other. MCR finds the path in the roadmap with the fewest geometric constraints or obstacles that need to be removed to connect the start state with the goal state. However, MCR does not direct where the obstacles should be moved. By encoding the obstacles with non-zero overlap to vanish our approach thus generalizes the MCR problem.
 
Navigation Among Movable Obstacles (NAMO)~\cite{stilman2005IJHR,nieuwenhuisen2008WAFR,van2009WAFR} is a related class of problems. NAMO is proved to be NP-hard~\cite{wilfong1988ASCS} and most approaches solve a subclass of problems that selects a set of obstacles to be moved and displaces them to reconfigure the environment. Another related area is the field of task and motion planning~\cite{kaelbling2013IJRR,srivastava2014ICRA,dantam2016RSS,garrett2018IJRR,thomas2020STAIRS,thomas2021RAS}. Selecting the obstacles to be displace may be accomplished via task planning and the geometric constraints may be incorporated into the symbolic planner~\cite{srivastava2014ICRA} or both the task and motion planning could be integrated by means of an efficient mapping between the two domains~\cite{dantam2016RSS,thomas2021RAS}. Manipulation among clutter or rearrangement planning in clutter~\cite{stilman2007ICRA,dogar2011RSS,krontiris2015RSS,karami2021arxiv} also bears resemblance to the problem proposed herein.~\cite{dogar2011RSS,karami2021arxiv} are often suboptimal in the sense that many objects may be moved than that is necessary due to their planner heuristic. 

%% file: conclusion.tex
We have presented an approach for minimum displacement motion planning that minimizes obstacles displacements to find a feasible path. The proposed approach is a two stage process where we first find a path with minimum robot-obstacle overlaps. A metric is used to measure the extant of overlap and this in turn is used to incrementally displace the obstacles to achieve the required feasible path. By appropriately tuning the weights $M_x$, $M_i$, $M_g$, the planning phase can be made to return the optimal solution. However, the refinement phase may not always return the optimal solution. For example an ideal displacement of an obstacle could be a combination of translation and rotation. However, currently we assume that the obstacle can either only translate or only rotate and therefore may render the refinement phase suboptimal. We plan to relax this assumption in future work. We also plan to generalize this work by doing away with bounding volume spheres to use the actual polyhedral geometry while computing the robot-obstacle interactions. 